\documentclass[letterpaper]{article}
\usepackage[margin=1in]{geometry}

\usepackage{times}
\usepackage{amsfonts} 
\usepackage{bm}       
\usepackage{amsmath}  

\usepackage{wrapfig}  

\usepackage{natbib}
\renewcommand{\cite}{\citep}

\usepackage{authblk} 

\usepackage{subcaption}
\usepackage{graphicx}
\graphicspath{{figures/}}
\usepackage{pdfpages}       

\usepackage[breaklinks]{hyperref}     

\usepackage{amsthm}
\newtheorem{lemma}{Lemma}

\newtheorem{theorem}{Theorem}

\title{Variational Inference for Gaussian Process with Panel Count Data}

\author[1]{Hongyi Ding}
\author[2]{Young Lee}
\author[1,3]{Issei Sato}
\author[3,1]{Masashi Sugiyama}
\affil[1]{The University of Tokyo, Japan}
\affil[2]{National University of Singapore, Singapore}
\affil[3]{The RIKEN Center for AIP, Tokyo, Japan}

\begin{document}

\maketitle

\begin{abstract}
We present the first framework for Gaussian-process-modulated Poisson processes when the temporal data appear in the form of panel counts. Panel count data frequently arise when experimental subjects are observed only at discrete time points and only the numbers of occurrences of the events between subsequent observation times are available. The exact occurrence timestamps of the events are unknown. The method of conducting the efficient variational inference is presented, based on the assumption of a Gaussian-process-modulated intensity function. We derive a tractable lower bound to alleviate the problems of the intractable evidence lower bound inherent in the variational inference framework. Our algorithm outperforms classical methods on both synthetic and three real panel count sets. 
\end{abstract}

\section{INTRODUCTION}
\paragraph{Background and issues.}
Temporal data frequently arise as outcomes of an underlying \emph{temporal point process} \cite{kingman1993poisson} in continuous time. Temporal data can generally be classified into two types. One is from experiments that monitor subjects in a continuous fashion; and thereby the exact timestamps of all occurrences of the events are fully observable. These data are usually referred to as \emph{recurrent event data} \cite{cook2007statistical}. On the other hand, we have the so-called \emph{panel count data} \cite{sun2016statistical}, which is the focus of our paper. Under this framework, subjects are examined or observed only at discrete time-points and thus give only the numbers of occurrences of the events between subsequent observation times. 

\begin{figure}
	\begin{subfigure}{.5\textwidth}
		\includegraphics[width=\columnwidth]{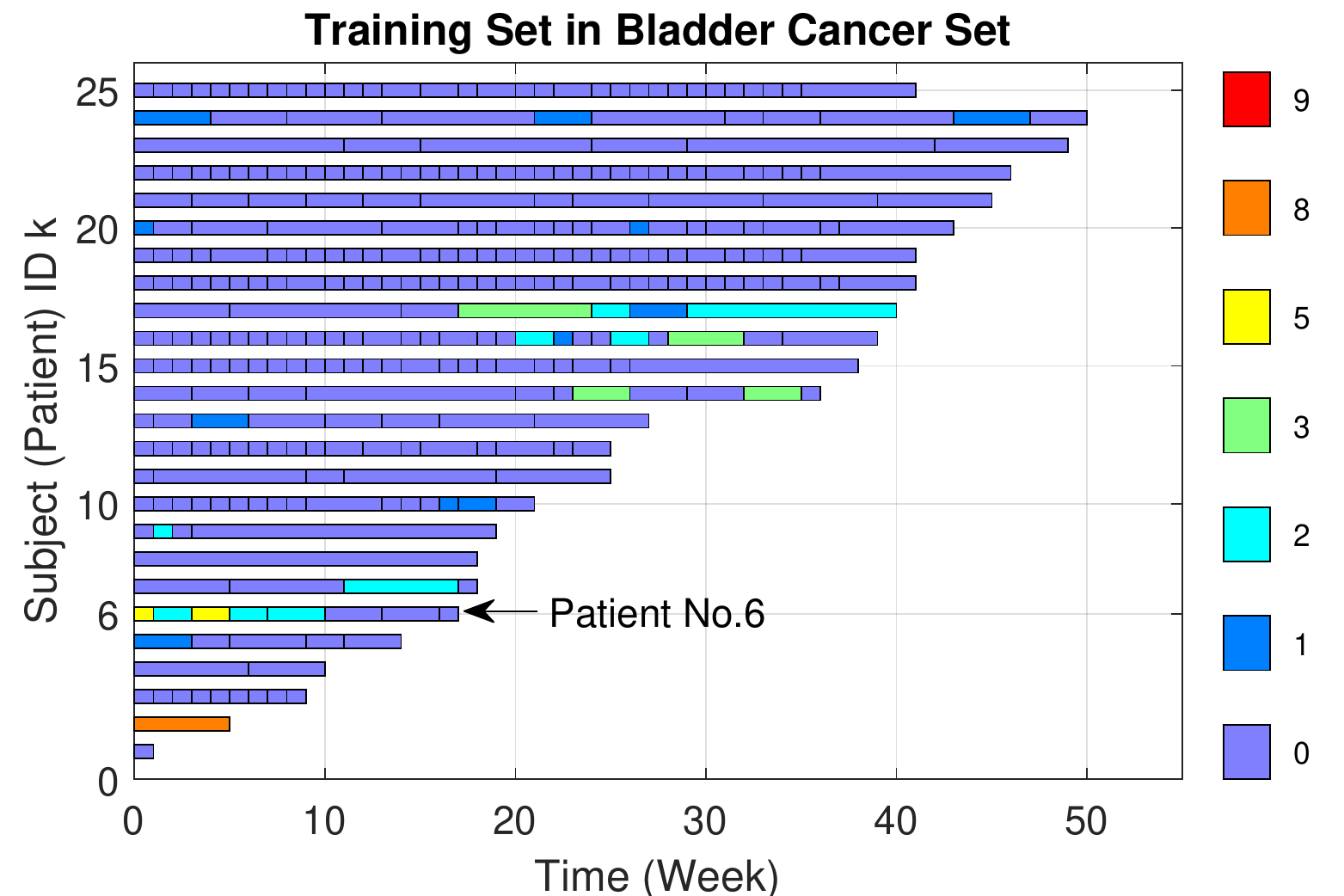}
		\caption{Illustration of the panel count data on the patients.}
		\label{Fig:DemoPanel}
	\end{subfigure}%
	\begin{subfigure}{.5\textwidth}
		\includegraphics[width=\columnwidth]{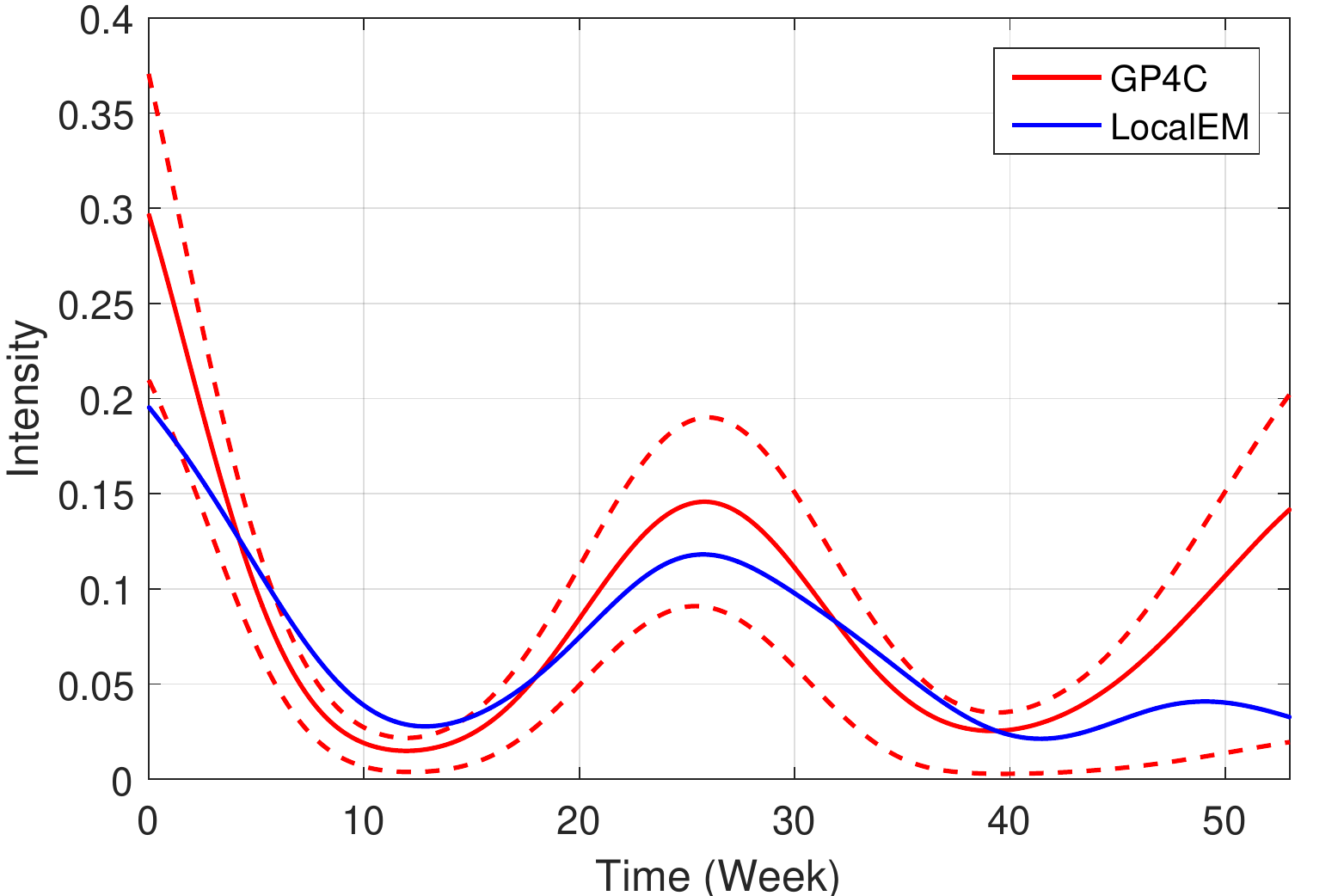}
		\caption{Inferred intensity function by the LocalEM and GP4C methods.}
		\label{Fig:DemoIntensity}
	\end{subfigure}
    \caption{(a) \textbf{Bladder Cancer Data Set}. For the $k$th subject (or the $k$th patient), his/her observation window $\mathcal{X}^{(k)}$ is divided into disjoint intervals. The $i$th interval is denoted as $\mathcal{X}_i^{(k)}$. For example, patient No. $6$ ($k=6$) has an observation window which is divided into 8 disjoint intervals, i.e., $\bigcup_{i=1}^8\mathcal{X}_i^{(6)}=\mathcal{X}^{(6)}$. Patients may drop out from the study at any time and this is the reason why the observation windows are different. An interval is shown by a rectangle. We use different colors to indicate the different numbers of new bladder tumors observed in this interval. Note that we only have access to the \emph{counts} in each interval. Our main aim is to infer the underlying intensity function in the panel count data. (b) \textbf{Bladder Cancer Data Set}. For GP4C, a 75\% credible interval is given by dotted lines. Our estimator GP4C provides the additional uncertainty in the estimated intensity function compared with LocalEM. See Section \ref{Sec:5} for details.}
\end{figure}

\paragraph{Characteristics of panel count data.} A common characteristic of the panel count data is that we only have the numbers of occurrences between subsequent observation times. In particular, the exact occurrence times of the events are unknown. Hence, panel counts are non-negative integers and they represent the number of occurrences of events within a fixed period. Classical examples often arise in the clinical trials \cite{thall1988analysis} where patients are required to go back to the hospital after a certain treatment and only the number of symptoms between subsequent visits are recorded, such as the number of vomits or new tumors. Figure \ref{Fig:DemoPanel} gives an example of panel count data. 

\paragraph{Objective of this study.}
The purpose of this paper is to present the variational Bayesian inference on \textbf{G}aussian-\textbf{p}rocess-modulated \textbf{P}oisson \textbf{p}rocesses (GP3) that permits panel data observations. 

There have been extensive studies on GP3 models and various inference algorithms are introduced for \emph{recurrent event data} when timestamps of the events are fully observable, e.g., Monte Carlo \cite{Diggle:2013,adams2009tractable}, Laplace approximation \cite{flaxman2015fast} and variational inference \cite{lloyd2015variational}. Among these approaches, the variational inference method \cite{lloyd2015variational} provides an efficient estimate of the intensity function and does not require a careful discretization of the underlying space.

To the best of our knowledge, however, there has not been any study carried out on the variational inference of the GP3 model when the data come in the form of panel counts. Our ultimate goal is to infer the underlying intensity function in the panel count data.


\paragraph{Related statistical works.} Based on the maximum likelihood criterion, several non-parametric estimators have been proposed to infer the underlying intensity function \cite{sun2016statistical}, e.g., a non-parametric maximum pseudo-likelihood estimator (NPMPLE) \cite{wellner2000two}, a non-parametric maximum pseudo-likelihood estimator with gamma frailty (NPMPLGF) \cite{zhang2003gamma} and the local Expectation-Maximization (LocalEM) estimator \cite{fan2011local}. Unlike NPMPLE and NPMPLGF, which only estimate the cumulative intensity function at a set of points, LocalEM provides a smooth estimate of the underlying intensity function due to the use of an exponential quadratic kernel \cite{fan2011local}. 

Besides the computational cost in selecting the bandwidth of the exponential quadratic kernel, the estimators obtained by the LocalEM algorithm and other similar algorithms are point-estimates in the sense that the estimated intensity function is a point in the functional space. These point estimates fail to capture the uncertainty in the data set. We show an example of the estimated intensity function by LocalEM in Figure \ref{Fig:DemoIntensity}. The uncertainty of the intensity function helps us understand the difficulty of the prediction at a given time.

\paragraph{Contributions.}
The contributions of our work are two-fold.
1) In the first place it undertakes to construct a variational inference procedure for \textbf{G}aussian-\textbf{P}rocess-modulated \textbf{P}oisson \textbf{P}rocess model for \textbf{P}anel \textbf{C}ount data (GP4C).  2) To carry out a variational inference in this setting, we derive a simple and tractable lower bound of the intractable evidence lower bound and demonstrate through empirical evidence that with this lower bound, GP4C outperforms the non-Bayesian method.


\section{BACKGROUND}
\label{Sec:2}
Throughout this paper, we denote the set of panel count data from $K\in \mathbb{N}^+$ independent subjects as $\mathcal{D}$. Each subject will generate a sequence of events in the continuous space $\mathcal{X}$. We only consider the temporal point processes where the continuous space $\mathcal{X}$ is a subset of $\mathbb{R}$.  In the \emph{recurrent event data}, the timestamps of the events are fully observable. We denote the timestamps from the $k$th subject as $\{x^{(k)}_j\in\mathcal{X}\}$. 

In the \emph{panel count data}, the $k$th subject is assessed in $N_k$ disjoint intervals $\{\mathcal{X}^{(k)}_i\}_{i=1}^{N_k}$, where $\cup_{i}\mathcal{X}^{(k)}_i=\mathcal{X}^{(k)}\subset \mathcal{X}$. We have access to each interval $\mathcal{X}^{(k)}_i$ and the number of events observed in this interval $m^{(k)}_i=|\{x^{(k)}_j\in \mathcal{X}^{(k)}_i\}|$. Let $\bm{d}_k=\{(\mathcal{X}^{(k)}_i,m^{(k)}_i)\}_{i=1}^{N_k}$ and $\mathcal{D}=\{\bm{d}_k\}$. Figure \ref{Fig:DemoPanel} illustrates an example of the panel count data. 

\subsection{LIKELIHOOD OF PANEL COUNT DATA}
In the \emph{recurrent event data}, one approach to modeling the events $\{x^{(k)}_j\in\mathcal{X}\}$ from each subject is to use the inhomogeneous Poisson processes (IPP) \cite{kingman1993poisson} and assume that there is a fixed underlying intensity function $\lambda(x): \mathcal{X}\rightarrow \mathbb{R}^+$. Given the intensity function $\lambda(x)$, the likelihood for observed events is 
\begin{equation}
p(\{x^{(k)}_j\}|\lambda(x))=\exp\Big(-\int_{\mathcal{X}}\lambda(x)dx\Big)\prod_{j}\lambda(x_j^{(k)}).
\end{equation}
To derive the likelihood of the \emph{panel count data} $\mathcal{D}$, we use two important features of an IPP \cite{kingman1993poisson}. The first is that given the intensity function $\lambda(x)$, the probability that we observe $m_i^{(k)}$ events in the interval $\mathcal{X}^{(k)}_i$ is given as follows:
\begin{equation}
p(m^{(k)}_i|\lambda(x);\mathcal{X}^{(k)}_i)=\frac{r_{ik}^{m^{(k)}_i}}{m^{(k)}_i!}\exp(-r_{ik}),
\label{Equ:Feat1}
\end{equation}
where $r_{ik}\stackrel{\Delta}{=}\int_{\mathcal{X}^{(k)}_i}\lambda(x)dx$ is the rate parameter of a Poisson distribution. Hereafter, we omit the dependency on $\mathcal{X}^{(k)}_i$ for simplicity. The second feature is that on two disjoint intervals $\mathcal{X}^{(k)}_i$ and $\mathcal{X}^{(k)}_j$ ( $\mathcal{X}^{(k)}_i\bigcap\mathcal{X}^{(k)}_j=\emptyset$), the number of events on these intervals are independent random variables.
\begin{equation}
p(m^{(k)}_j,m^{(k)}_i|\lambda(x))=p(m^{(k)}_j|\lambda(x))p(m^{(k)}_i|\lambda(x)).
\label{Equ:Feat2}
\end{equation}
Based on these two features, the likelihood of the panel count data $\mathcal{D}$ can be derived. We assume that all subjects share the same intensity function $\lambda(x)$. Since $K$ subjects are independent of each other and for the $k$th subject, the $N_k$ intervals $\{\mathcal{X}^{(k)}_i\}_{i=1}^{N_k}$ are disjoint, we obtain the following likelihood
\begin{equation}
p(\mathcal{D}|\lambda(x))=\prod_{k=1}^{K}p(\bm{d}_k|\lambda(x))=\prod_{k=1}^{K}\prod_{i=1}^{N_k}p(m^{(k)}_i|\lambda(x)).
\label{Equ:likelihood}
\end{equation}
Several maximum likelihood estimators have been proposed on the basis of this likelihood or its variants, e.g., NPMPLE \cite{wellner2000two,wellner2007two}, NPMPLGF \cite{zhang2003gamma} and the LocalEM estimator \cite{fan2011local}. An estimate from LocalEM on the data set in Figure \ref{Fig:DemoPanel} is given in Figure \ref{Fig:DemoIntensity}. As we discussed, these estimators fail to model the uncertainty in the intensity function.


\subsection{GP3 MODEL}
In order to model the uncertainty of the intensity function $\lambda(x)$ via a kernel, the traditional approach is to use the Cox process \cite{kingman1993poisson}. A Cox process is defined via a stochastic intensity function $\lambda(x)$. The stochastic process to generate the intensity function is usually chosen to be a Gaussian process (GP) \cite{adams2009tractable} and the model is called GP3 model.

For the \emph{recurrent event data}, GP3 models have been studied extensively \cite{adams2009tractable,gunter2014efficient,lloyd2015variational}. The following model is an example of GP3 models \cite{lloyd2015variational},
\begin{equation}
\lambda(x) = f^2(x),~ f\sim \mathcal{GP}(g(x),\kappa(x,x')),
\label{Equ:BasicModel}
\end{equation} 
where $\mathcal{GP}(g(x),\kappa(x,x'))$ denotes the Gaussian process with mean function $g(x)$ and covariance function $\kappa(x,x')$. The function $f(x)$ drawn from a GP prior is squared to ensure the non-negativity of the intensity function. The GP3 model in Equation \eqref{Equ:BasicModel} admits a complete variational inference framework. Moreover, this intensity model can be enhanced with an independent variable for each subject or a mixture structure \cite{lloyd2016latent} to flexibly model the heterogeneity of the intensity functions across several subjects.

\section{OUR MODEL GP4C : GP3 MODEL FOR PANEL COUNT DATA}
In order to retain the scalability and efficiency of the variational inference approach \cite{lloyd2015variational} and add the uncertainty on the intensity function when we only observe the panel count data, we use the GP3 model defined in Equation \eqref{Equ:BasicModel} as the underlying intensity model. 

The joint distribution $p(\mathcal{D},f)$ can be obtained by combining the likelihood model in Equation \eqref{Equ:likelihood} and the intensity model in Equation \eqref{Equ:BasicModel}.
\begin{equation}
p(\mathcal{D},f) = \Big[\prod_{k=1}^{K}p(\bm{d}_k|\lambda(x))\Big]p(f;g,\kappa).
\end{equation}
We call this model \textbf{GP}-modulated \textbf{P}oisson \textbf{P}rocess model for \textbf{P}anel \textbf{C}ount data (GP4C).

\section{INFERENCE}
In this section, we will discuss the problems when applying variational inference techniques on the GP4C model.
\subsection{VARIATIONAL INFERENCE}
We use sparse GPs to reduce the computational complexity with the set of pseudo inputs $\{x_r\}_{r=1}^R$ on $\mathcal{X}$ \cite{titsias2009variational}. Let $\bm{f}_R\stackrel{\Delta}{=}[f(x_1),\ldots,f(x_R)]^\top$. The joint model with additional pseudo inputs is $p(\mathcal{D},f,\bm{f}_R)=p(\mathcal{D}|f)p(f|\bm{f}_R)p(\bm{f}_R)$ and the variational distribution is defined as follows:
\begin{equation}
q(f,\bm{f}_R)=p(f|\bm{f}_R)q(\bm{f}_R), 
\label{Equ:VarDis}
\end{equation}
where $q(\bm{f}_R)= \mathcal{N}(\bm{\mu},\Sigma)$ and $\mathcal{N}(\bm{\mu},\Sigma)$ denotes the normal distribution with mean $\bm{\mu}$ and covariance matrix $\Sigma$. The evidence lower bound (ELBO) $\mathcal{L}$ can be obtained by using Jensen's inequality.
\begin{align}
&\ln p(\mathcal{D})  \geq \iint q(f,\bm{f}_R)\ln \frac{p(\mathcal{D},f,\bm{f}_R)}{q(f,\bm{f}_R)} dfd\bm{f}_R\nonumber\\
& = \sum_{k=1}^{K}\sum_{i=1}^{N_k}\Big(m_i^{(k)}\mathbb{E}_q\Big[\ln\int_{\mathcal{X}^{(k)}_i}f^2(x)dx\Big]-\ln (m_i^{(k)}!)\Big)- \sum_{k=1}^{K}\mathbb{E}_q\Big[\int_{\mathcal{X}^{(k)}}f^2(x)dx\Big] +\mathbb{E}_q\Big[\ln \frac{p(\bm{f}_R)}{q(\bm{f}_R)}\Big]\stackrel{\Delta}{=}\mathcal{L}.
\label{Equ:ELBO}
\end{align}
In ELBO, when assuming that the covariance function $\kappa(x,x')$ is the automatic relevance determination (ARD) function $\kappa(x,x')=\gamma\exp\Big(-\frac{(x-x^\prime)^2}{2a^2}\Big)$, $x, x^\prime\in \mathcal{X}$, the second term in the ELBO can be analytically calculated \cite{lloyd2015variational} as follows:
\begin{equation}
\mathbb{E}_q\Big[\int_{\mathcal{X}^{(k)}}f^2(x)dx\Big]=\gamma|\mathcal{X}^{(k)}|-\mathrm{tr}(K_{RR}^{-1}\Phi)+\mathrm{tr}(K_{RR}^{-1}\Phi K_{RR}^{-1}(\bm{\mu\mu}^\top+\Sigma)),
\label{Equ:Integral}
\end{equation}
where $\Phi$ is an $R\times R$ matrix related to the pseudo inputs with its $i,j$'th entry equal to $\int_{\mathcal{X}^{(k)}}\kappa(x_i,x)\kappa(x,x_j)dx$ and $K_{RR}$ is the covariance matrix computed at the pseudo inputs. However, the ELBO $\mathcal{L}$ is still intractable, since we can not analytically compute the expected integral $\mathbb{E}_q\Big[\ln\int_{\mathcal{X}^{(k)}_i}f^2(x)dx\Big]$ in the first term. 

\subsection{A TRACTABLE LOWER BOUND}
We tackle the intractable expectation by deriving a tractable lower bound. First we introduce a relevant lemma on the expectation of the logarithm of the square of a normal-distributed random variable. 
\begin{lemma}
	Let $y\sim \mathcal{N}(\mu,\sigma^2)$ and $\varphi = (\mu/\sigma)^2$. Then 
	\begin{equation}
	\mathbb{E}_y[\ln y^2] = \ln(2\sigma^2)+\sum_{j=0}^{\infty} \frac{(\varphi/2)^j\exp(-\varphi/2)}{j!}\psi(j+1/2),
	\end{equation}
	where $\psi(\cdot)$ is the digamma function.
	\label{lem:1}	
\end{lemma}
The proof of Lemma \ref{lem:1} can be found in Appendix A. Let 
\begin{equation}
g_m(y) = \sum_{j=0}^{\infty} \frac{y^j\exp(-y)}{j!}\psi(j+m). 
\end{equation}
Then $\mathbb{E}_y[\ln y^2]=\ln(2\sigma^2)+g_{0.5}(\varphi/2)$. The function $g_m(y)$, where $y$ is a positive real number and $m$ is a positive integer, has been studied in the analysis of mobile and wireless communication systems \cite{moser2007some}. For $m=1/2$, $g_{0.5}(\varphi/2)$ can be computed using a hyper-geometric confluent function $G(\cdot)$ \cite{lloyd2015variational}, which is stored in a pre-computed look-up table. 
\begin{equation}
g_{0.5}(\varphi/2)=-G(-\varphi/2)-2\ln 2 - C,
\label{Equ:TrueG}
\end{equation}
where $C$ is Euler's constant and $C\approx 0.5772$. However, to the best of our knowledge, it is still not clear how to calculate the integral of the function $G(-\varphi/2)$ when using a GP. To derive a tractable lower bound of the intractable expectation, we introduce the following lemma to give a lower bound of the function $g_m(y)$ and the proof can be found in Appendix B.

\begin{lemma}
	Let $y\sim \mathcal{N}(\mu,\sigma^2)$ and $C$ be Euler's constant.
	\begin{equation}
	\mathbb{E}_y[\ln y^2]\geq \ln(\mu^2+b\sigma^2)-C-\ln 2,~\forall b\in[0,1].
	\end{equation}
	\label{lem:2}
\end{lemma}

Based on Lemma \ref{lem:2}, we compute a lower bound for the intractable expectation in the ELBO.

\begin{theorem}
	Let $f$ be a GP as defined in equation (\ref{Equ:BasicModel}). For $b\in [0,1]$, the following bound holds:
	\begin{equation}
	\mathbb{E}_q\Big[\ln\int_{\mathcal{X}^{(k)}_i}f^2(x)dx\Big]\geq-C-\ln 2 +\ln\Big(\int_{\mathcal{X}^{(k)}_i}\Big(\mathbb{E}_q^2f(x)+b\mathrm{Var}_qf(x)\Big)dx\Big),
	\end{equation}
	where the distribution $q$ is given in Equation \eqref{Equ:VarDis}.
	\begin{proof}
		We first use Jensen's inequality on the logarithm function and then interchange the order of integration and expectation.
		\begin{equation}
		\mathbb{E}_q\Big[\ln\int_{\mathcal{X}^{(k)}_i}f^2(x)dx\Big]=\mathbb{E}_q\Big[\ln\int_{\mathcal{X}^{(k)}_i}\tilde{p}(x)\frac{f^2(x)}{\tilde{p}(x)}dx\Big]\geq \int_{\mathcal{X}^{(k)}_i}\tilde{p}(x)\mathbb{E}_q\Big[\ln\frac{f^2(x)}{\tilde{p}(x)}\Big]dx,
		\label{Equ:lower}
		\end{equation}
		where $\tilde{p}(x)$ is a probability distribution on $\mathcal{X}^{(k)}_i$. Furthermore, maximizing this lower bound with respect to $\tilde{p}(x)$ yields the optimal distribution:
		\begin{equation}
		\tilde{p}_{\mathrm{opt}}(x) \propto \exp\Big(\mathbb{E}_q\ln f^2(x)\Big).
		\label{Equ:opt}
		\end{equation}
		We remark that this result is analogous to that of the discrete version presented in \citet{paisley2010two}. Substituting equation \eqref{Equ:opt} into the right-hand side of Equation \eqref{Equ:lower} yields
		\begin{align*}
		&\mathbb{E}_q\Big[\ln\int_{\mathcal{X}^{(k)}_i}f^2(x)dx\Big]\geq \ln\Big(\int_{\mathcal{X}^{(k)}_i}e^{\mathbb{E}_q\ln f^2(x)}dx\Big)\nonumber\\
		& \stackrel{(13)}{\geq} \ln\Big(\int_{\mathcal{X}^{(k)}_i}e^{\ln(\mathbb{E}_q^2f(x)+b\mathrm{Var}_qf(x))-C-\ln 2}dx\Big)  = \ln\Big(\int_{\mathcal{X}^{(k)}_i}\Big(\mathbb{E}_q^2f(x)+b\mathrm{Var}_qf(x)\Big)dx\Big)-C-\ln 2.
		\end{align*}
		where we have invoked Lemma \ref{lem:2} in the penultimate line whilst defining $y:=f(x)$.
	\end{proof}
	\label{Theo:1}
\end{theorem}
It should be emphasized that we are making no further assumptions on the dimensionality of $x$ in the proof of Theorem \ref{Theo:1}. Hence we may augment the dimensionality of $x$ in Theorem \ref{Theo:1} such that it can also be applied to problems in spatial point processes. In summary, the ELBO in Equation \eqref{Equ:ELBO} inherits an analytical bound. We present the following:
\begin{theorem}
	A tractable lower bound of the ELBO $\mathcal{L}$ in the GP4C model is given as follows:
	\begin{align}
	\mathcal{L}\geq\mathcal{\tilde{L}}\stackrel{\Delta}{=}&- \sum_{k=1}^{K}\mathbb{E}_q\Big[\int_{\mathcal{X}^{(k)}}f^2(x)dx\Big] +\mathbb{E}_q\Big[\ln \frac{p(\bm{f}_R)}{q(\bm{f}_R)}\Big] + \sum_{k=1}^{K}\sum_{i=1}^{N_k}m_i^{(k)}\ln\Big(\int_{\mathcal{X}^{(k)}_i}\Big(\mathbb{E}_q^2f(x)+b\mathrm{Var}_qf(x)\Big)dx\Big)\nonumber\\
	&-\sum_{k=1}^{K}\sum_{i=1}^{N_k}\Big(m_i^{(k)}(C+\ln 2)+\ln (m_i^{(k)}!)\Big).
	\label{Equ:ModiELBO}
	\end{align}
\end{theorem}
The details of the proof are deferred to Appendix C. The derivations of $\mathbb{E}_q^2f(x)$ and $\mathrm{Var}_qf(x)$ follow similar lines to the derivation of Equation \eqref{Equ:Integral}. The third part of $\mathcal{\tilde{L}}$ is a constant and thus can be omitted when maximizing the lower bound. Let $\Psi=\{\bm{\mu},\Sigma\}$ and $\Phi=\{\gamma,a\}$ be the variational parameters and hyper-parameters in the covariance function of a GP, respectively. We use the variational Expectation-Maximization (vEM) algorithm \cite{dempster1977maximum} to update the parameters $\Psi$ and $\Phi$ iteratively on the modified ELBO $\mathcal{\tilde{L}}$. 

\subsection{THE VALUE OF PARAMETER $b$}
\label{Sec:choiceB}
A natural question is, how do we select the parameter $b$ in Theorem \ref{Theo:1}? Recall that two inequalities were used in the proof. For inequality \eqref{Equ:lower}, it is cumbersome to evaluate since it is an integral over $\mathcal{X}_i^{(k)}$. We first examine different choices of $b$ in Lemma \ref{lem:2}. 

\begin{figure}[t]
	\centering
	\includegraphics[width=0.7\columnwidth]{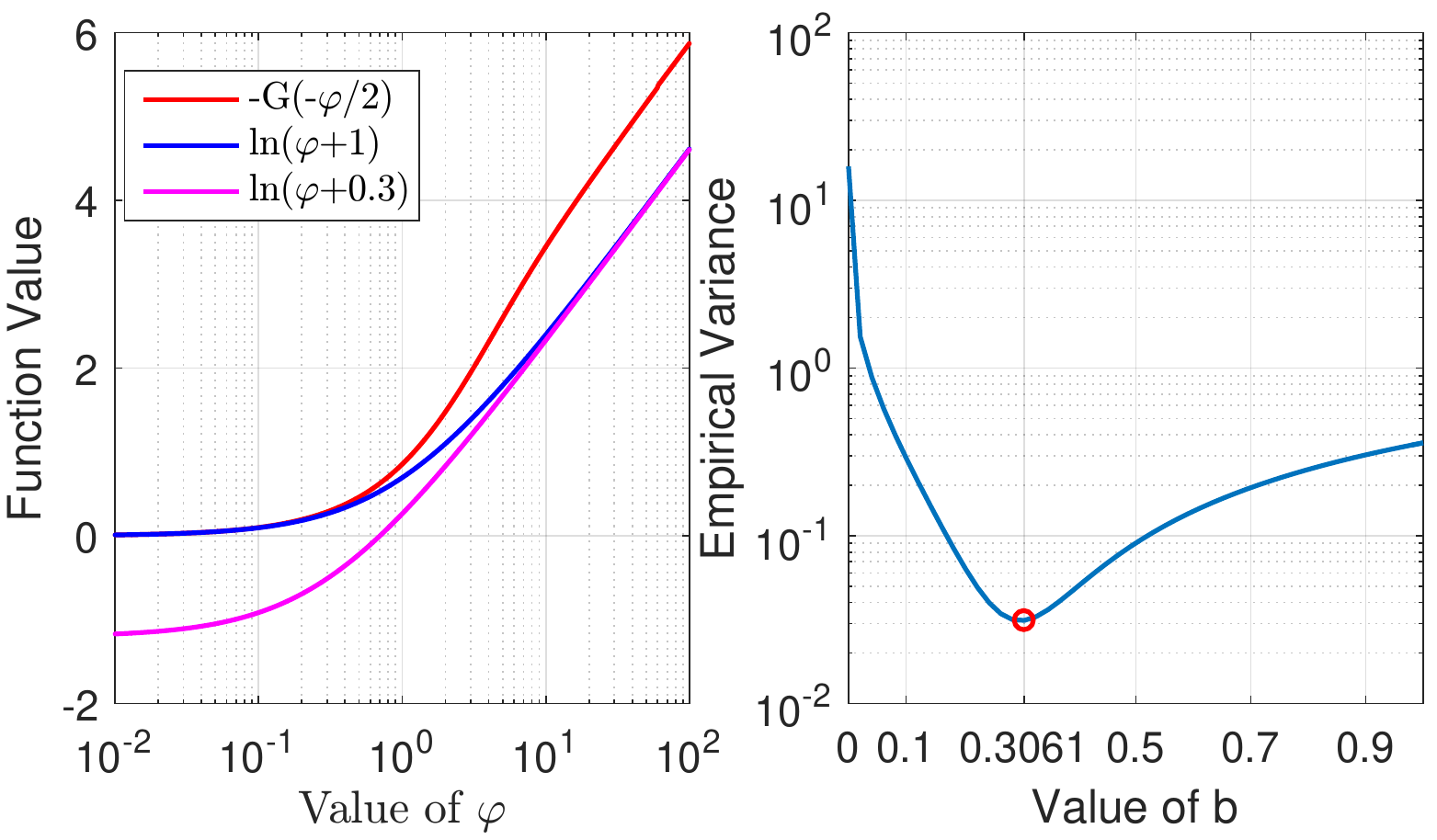}
	\caption{\textbf{Influences of $b$ in Lemma 2}. (Left) The true value of $-G(-\varphi/2)$ by a look-up table and two simple lower bounds. The bound $\ln(\phi+b)$ with $b=0.3$ correlates with the curve of the true value better. (Right). The variance $\mathrm{Var}[h(\varphi,b)]$ when varying the choices of $b$ and the best $b$ is shown with a red circle.}
	\label{Fig:HyperBound}
\end{figure}

In \citet{paisley2012variational}, a more correlated lower bound of the ELBO serves as a better control variate in reducing the variance of a stochastic gradient. Inspired by this study, we introduce a heuristic method and conduct the following experiment to evaluate the correlation for different choices of $b$. In Lemma \ref{lem:2}, the difference between the lower bound and the true value is 
\begin{equation}
\ln(\mu^2+b\sigma^2)-C-\ln 2-\mathbb{E}_y[\ln y^2] = \ln (\varphi+b)+G(-\varphi/2)\stackrel{\Delta}{=}h(\varphi,b).
\end{equation}
We vary $\varphi = (\mu/\sigma)^2$ on a vector of 5000 logarithmically spaced points between $10^{-6}$ and $10^{6}$ and evaluate the correlation between the lower bound and the true value by the variance of the difference $\mathrm{Var}[h(\varphi,b)]$. We calculate $\mathrm{Var}[h(\varphi,b)]$ on a vector of 50 evenly spaced choices of $b$ between 0 and 1 and the result is shown in Figure \ref{Fig:HyperBound}. We see that the optimal choice of $b$ is 0.3061 if $\varphi$ ranges from $10^{-6}$ to $10^{6}$. In the actual situation, this optimal value of $b$ depends on the range of the $\varphi$ in the data and the influence of Inequality \eqref{Equ:lower}, we evaluate several choices of $b$ on synthetic data sets in Section \ref{Sec:5}.

\subsection{COMPUTATIONAL COMPLEXITY}
\label{Sec:Comp}
Let each interval in temporal point processes be $\mathcal{X}_i^{(k)}=[x_{ai}^{(k)},x_{bi}^{(k)}]$ with two end points $x_{ai}^{(k)}$ and $x_{bi}^{(k)}$. Two intervals are different if at least one end point is different. We denote the number of different intervals in the data set as $N$ and the number of pseudo inputs as $M$. For each interval, the computation complexity of GP4C is $\mathcal{O}(M^3)$ which is determined by the matrix-matrix calculation when evaluating the $\mathrm{Var}_qf(x)$ in Equation \eqref{Equ:ModiELBO}. The computational complexity during one iteration of the vEM algorithm is $\mathcal{O}(NM^3)$ since in our implementation, we calculate the integral of all $N$ different intervals.

We analyze the computational complexity of the LocalEM \cite{fan2011local} algorithm for comparison.  In LocalEM, $\{x_{ai}^{(k)}\}$ and $\{x_{bi}^{(k)}\}$ are first merged into a single ordered set $X$ where duplicated values are removed. We denote the size of the merged set $X$ as $\bar{N}$ and generally $\bar{N}\leq N$. Then the Gaussian quadratic rule with $\bar{M}$ points is used to calculate the integral of the intensity function between subsequent values in the set $X$ and the computational complexity during one iteration is $\mathcal{O}(\bar{N}^2\bar{M}^2)$. If the size of merged set $\bar{N}$ is significantly smaller than $N$, LocalEM may be computationally more efficient than GP4C. However, if $\bar{N}\approx N$, LocalEM may suffer from the term $\bar{N}^2$ in the computational complexity.

\section{EXPERIMENTS}
\label{Sec:5}
We evaluate our proposed GP4C model and compare it with the benchmark methods on both synthetic and real-world data sets.
For each data set $\mathcal{D}$, we randomly partition the subjects into training and testing sets, which we denote as $\mathcal{D}_\mathrm{train}$ and $\mathcal{D}_\mathrm{test}$, respectively. We repeat each setting for $S=40$ times. In the $s$th trial, the training and testing sets are denoted as $\mathcal{D}_{\mathrm{train}}^{(s)}$ and $\mathcal{D}_{\mathrm{test}}^{(s)}$.

\textbf{Benchmark}. Two benchmark algorithms are used in our experiments.
\begin{itemize}
	\item [a)] We implemented GP3 using variational inference \cite{lloyd2015variational}. This benchmark reflects the best performance that can be obtained if we obtain the recurrent event data set where we have the exact timestamps.
	
	\item [b)] We use the LocalEM algorithm as another benchmark, since both LocalEM and GP4C are nonparametric estimators based on the maximum likelihood criterion. To fairly compare the computation time, we implemented the LocalEM algorithm in MATLAB based on the R code provided in \citet{fan2011local}. This method produces a smooth estimate of the intensity function due to the use of an exponential quadratic kernel. We use a 5-fold cross-validation on the training data set to select the bandwidth of the exponential quadratic kernel. 
\end{itemize}

\textbf{Evaluation Metric}. We evaluate the performance of the algorithms in terms of three metrics.
\begin{itemize}
	\item [a)]Mean of the integrated squared error ($\mathrm{MISE}$).  In synthetic data sets, we have the ground truth of the intensity function $\lambda_\mathrm{true}$ and the integrated squared error can be calculated using our estimated intensity function $\lambda_\mathrm{est}^{(s)}$ during the $s$th trial. To measure the bias of each estimator, we calculate the mean of the integrated squared error as follows:
	\begin{equation}
	\mathrm{MISE}(s)\stackrel{\Delta}{=}  \int_{\mathcal{X}}(\lambda^{(s)}_\mathrm{est}(x)-\lambda_\mathrm{true}(x))^2dx.
	\end{equation}
	For GP4C, to measure its bias, we omit the variance of the estimator and use the expectation of the intensity function $\mathbb{E}_{q^{(s)}}[f^2(x)]$ as $\lambda^{(s)}_\mathrm{est}(x)$ to calculate MISE.
	\item [b)]Test log likelihood $\mathcal{L}_\mathrm{test}$. During the $s$th trial, the logarithm of the test likelihood can be written as follows:
	\begin{equation}
	\mathcal{L}_\mathrm{test}(s) \stackrel{\Delta}{=} \ln \int p(\mathcal{D}_{\mathrm{test}}^{(s)}|f)p(f|\mathcal{D}_\mathrm{train}^{(s)}) df.
	\end{equation}
	In the actual computation, we omit the complicated constant term $\sum_{k}\sum_i\ln (m_i^{(k)}!)$ in $p(\mathcal{D}_{\mathrm{test}}^{(s)}|f)$. Due to this omission, we use only the average of the test likelihood $1/S\sum_{s}\mathcal{L}_\mathrm{test}^{(s)}$ to measure the quality of the estimator. For LocalEM, since this estimator provides a point estimate and we directly use the estimated function $f^{(s)}$ to calculate $\mathcal{L}_\mathrm{test}(s)$. For GP4C and GP3, we need to sample the function $f$ from the variational distribution and the detailed calculation can be found in Appendix D.
	\item [c)]Computation time $T$. We record the training time measured in seconds for each setting. For GP3 and GP4C, we record the computation time of the training process. For LocalEM, it includes the time of a 5-fold cross-validation on the training set to select the bandwidth of the exponential quadratic kernel and the time of a training process over the whole training set.
	
\end{itemize}

\textbf{Experiment Settings}. For GP3 and GP4C, following \citet{lian2015multitask}, we use the re-parametrization trick $\Sigma = LL^\top$ by Cholesky decomposition and add positivity constraints to the diagonal elements in $L$. Due to this constraint on $L$, we use the limited-memory projected quasi-Newton algorithm \cite{schmidt2009optimizing} to optimize the variational parameters $\Psi=\{\bm{\mu},\Sigma\}$. We add a jitter term $\epsilon I$ where $\epsilon = 10^{-6}$ to the covariance matrix $K_{RR}$ to avoid numerical instability \cite{titsias2009variational}. 

\subsection{SYNTHETIC DATA SETS}
We test three synthetic data sets which we denote as Synthetic A, B and C data sets, respectively. 

\begin{figure}
	\begin{subfigure}{.5\textwidth}
		\includegraphics[width=\columnwidth]{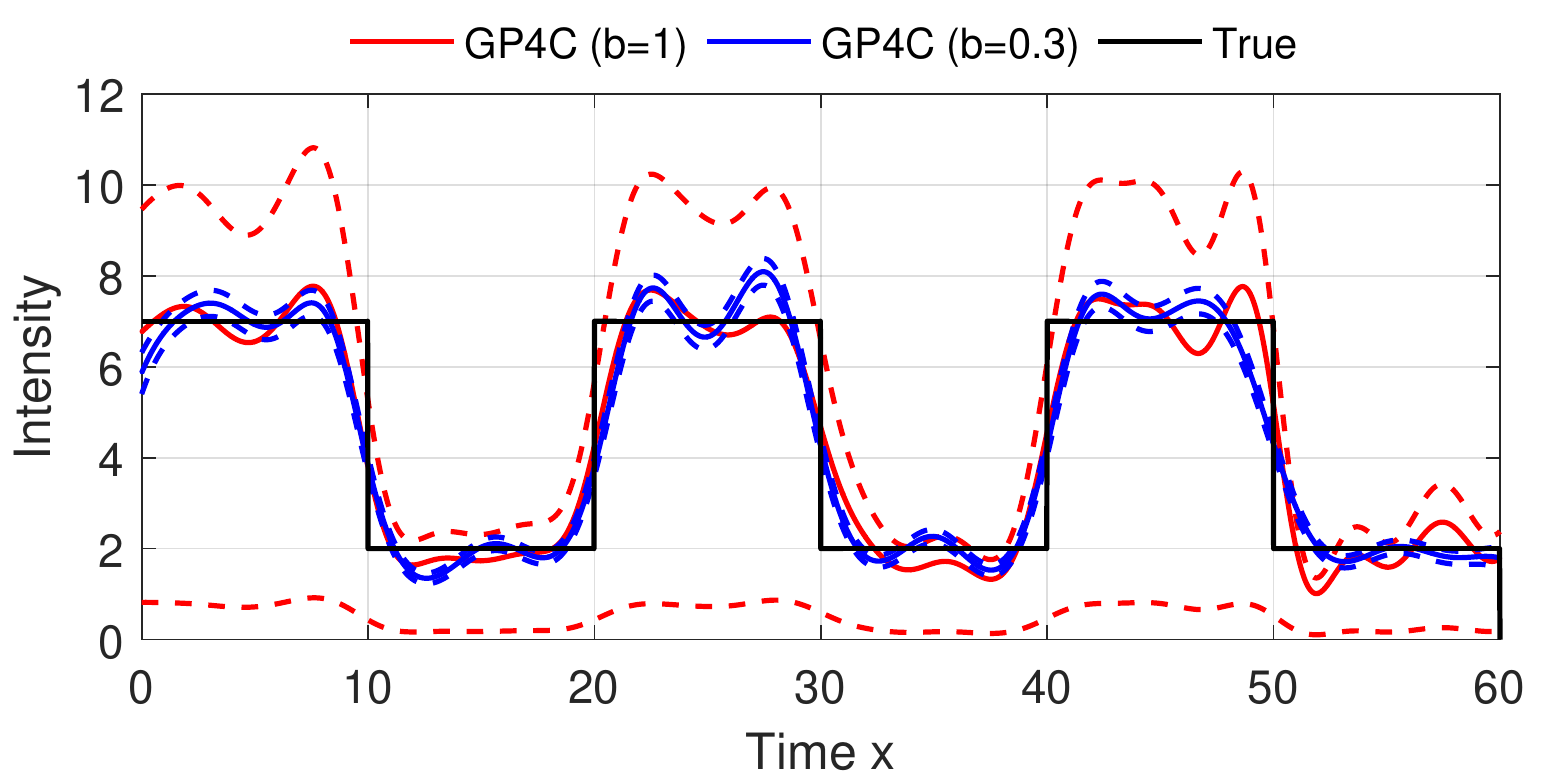}
		\caption{\textbf{Synthetic A Data Set}. }
		\label{Fig:BadB}
	\end{subfigure}%
	\begin{subfigure}{.5\textwidth}
		\includegraphics[width=\columnwidth]{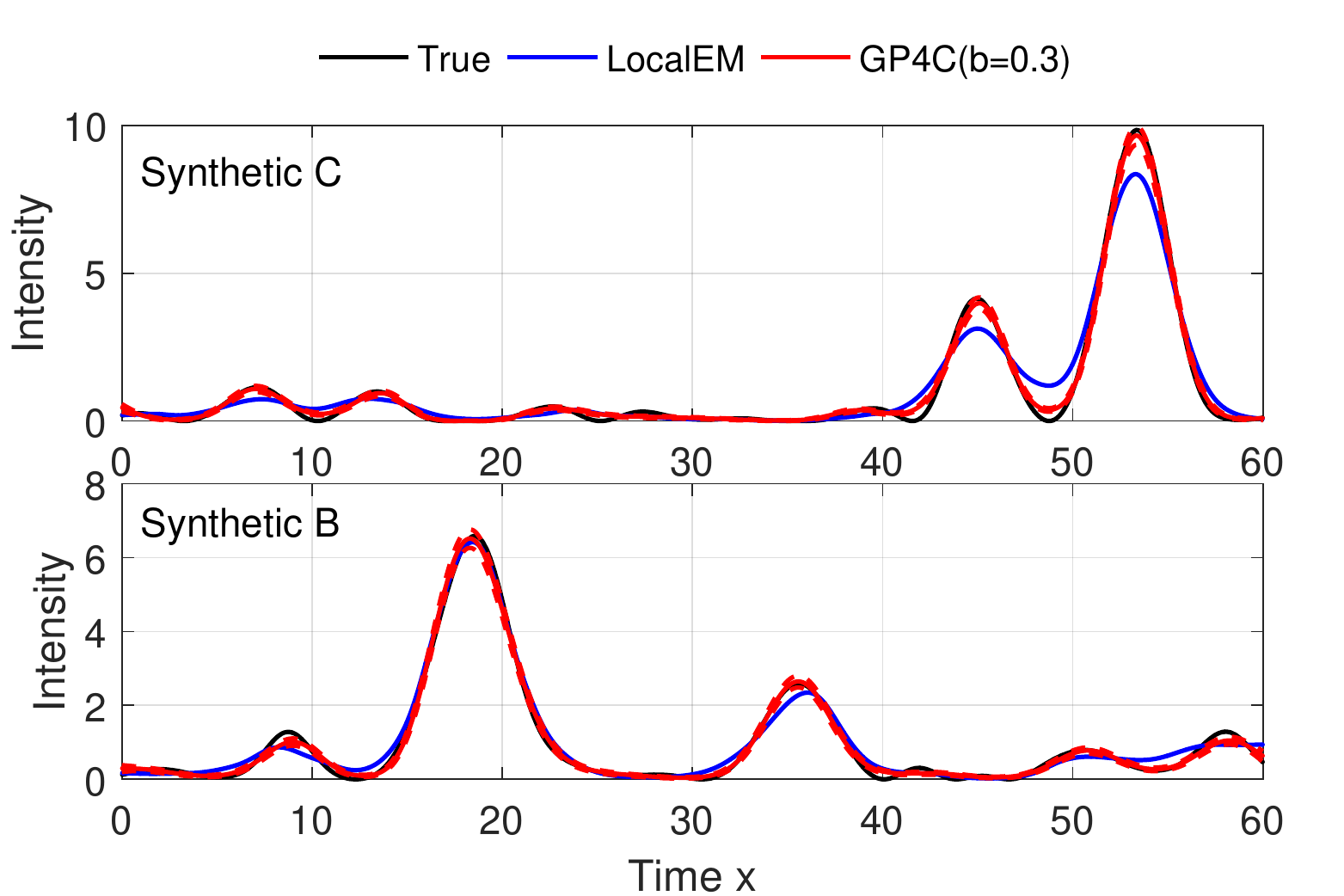}
		\caption{\textbf{Synthetic B \& C Data Sets}.}
		\label{Fig:SynBC}
	\end{subfigure}
	\caption{(a) The estimated intensity functions from GP4C ($b=1$) and GP4C ($b=0.3)$ are shown with 75\% credible intervals. True intensity function $h_1(x)$ is given for comparison. We see that GP4C ($b=1$) over-estimates the variance of the intensity function. (b) Inferred intensity function by the LocalEM and GP4C methods on Synthetic B and Synthetic C data sets. The underlying intensity function is drawn from a Gaussian process. For GP4C, a 75\% credible interval is given by dotted lines.}
\end{figure}

On Synthetic A data set, the intensity function is a square wave function $h_1(x)$ as follows. See Figure \ref{Fig:BadB} for an illustration of $h_1(x)$. 
\begin{align*}
h_1(x) & = \left \{ 
\begin{aligned}
7,~&\mathrm{if}~\mathrm{mod}\Big(\Big[\frac{x}{10}\Big],2\Big) = 0,\\
2,~&\mathrm{otherwise}.
\end{aligned}
\right.
\end{align*}
On Synthetic B and C data set, the underlying intensity functions are drawn according to Equation \eqref{Equ:BasicModel}. We first draw a function from a GP on a vector of 3001 evenly-spaced points in $\mathcal{X} = [0,T]$, where $T=60$. We approximate the value of the function at an arbitrary position with linear interpolation. The function is then squared to guarantee the positiveness of the intensity function. See Figure \ref{Fig:SynBC} for an illustration of the two intensity functions.

During the $s$th trial, we first generate a \emph{recurrent event data set} with 100 subjects on the same observation window $\mathcal{X}^{(k)}=\mathcal{X}$. Then we generate the corresponding \emph{panel count data set} $\mathcal{D}^{(s)}$ by censoring each subject with 10 intervals. We generate the censored intervals by a draw from a Dirichlet distribution $\bm{w}^{(k)}\sim\mathrm{Dir}(\bm{\theta})$ and $\bm{\theta}$ is a 10-dimensional vector with all elements equal to 1. The $i$th interval of the $k$th subject can be computed as $\mathcal{X}_i^{(k)}=[\sum_{j=1}^{i-1}w_j^{(k)}T,\sum_{j=1}^{i}w_j^{(k)}T]$. We randomly partition $\mathcal{D}^{(s)}$ into two parts, where 50 subjects are used for training and 50 for testing. 

\begin{table}[t]
	\caption{\textbf{Synthetic data sets}. Statistics about different choices of $b$. GP3 uses the \emph{recurrent event data} while LocalEM and GP4C use the \emph{panel count data}. The best performance among GP4C and LocalEM is marked with bold font. $b=0.3$ and $b=0$ perform better than $b=1$ in terms of MISE and $\mathcal{L}_\mathrm{test}$. }
	\label{Tab:Choicesb}
	\begin{center}
		\begin{tabular}{lllll}
			\multicolumn{1}{c}{\textbf{Method}} &\multicolumn{1}{c}{$\mathcal{L}_\mathrm{test}$}  &\multicolumn{1}{c}{\textbf{MISE}}& \multicolumn{1}{c}{$T [s]$}\\
			\hline \\[-1.5ex]
			(Synthetic A)&&&\\
			GP3 & 37099.66 & 29.23$\pm$1.27 & 12.70\\
			LocalEM & 37088.12 & 46.06$\pm$3.25 & 21.03\\
			GP4C ($b=0$) & 37091.24 & 41.72$\pm$3.38 & \textbf{20.20}\\
			GP4C ($b=0.3$) & \textbf{37091.67} & \textbf{41.18$\pm$3.61} & 22.89\\
			GP4C ($b=1$) & 35635.63 & 42.71$\pm$6.34 & 27.89\\
			\hline\\[-1.5ex]
			(Synthetic B)&&&\\
			GP3 & 4473.25 & 0.52$\pm$0.18 & 7.70\\
			LocalEM & 4454.53 & 4.70$\pm$1.43 & 21.94\\
			GP4C ($b=0$) & 4465.90 & \textbf{1.59$\pm$0.53} & 19.51\\
			GP4C ($b=0.3$) & \textbf{4466.17} & 1.63$\pm$0.52 & \textbf{18.70}\\
			GP4C ($b=1$) & 4184.74 & 2.17$\pm$0.83 & 32.23\\
			\hline\\[-1.5ex]
			(Synthetic C)&&&\\
			GP3 & 6213.15 & 1.06$\pm$0.37 & 8.52\\
			LocalEM & 6175.75 & 17.67$\pm$3.40 & 22.10\\
			GP4C ($b=0$) & 6204.30 & 2.38$\pm$0.76 & 16.77\\
			GP4C ($b=0.3$) & \textbf{6205.17} & \textbf{2.34$\pm$0.79} & \textbf{16.70}\\
			GP4C ($b=1$) & 5954.84 & 2.49$\pm$1.02 & 27.89\\
		\end{tabular}
	\end{center}
\end{table} 

\textbf{Different choices of the hyper-parameter $b$}. On all three synthetic data sets, we test three different choices of $b$ in $\{0,0.3,1\}$. We choose the number of pseudo inputs to be $30$. We calculate the MISE and $\mathcal{L}_\mathrm{test}$ and the results are provided in Table \ref{Tab:Choicesb}. We see that $b=0,0.3$ generally outperform $b=1$ on these simple synthetic data sets. However, the difference between $b=0$ and $b=0.3$ is not significant. The reason is that Inequality \eqref{Equ:lower} and the range of $\varphi$ on $\mathcal{X}$ are also relevant to the actual performance of different $b$, as we discussed in Section \ref{Sec:choiceB}. 

To investigate the reason behind the bad performance of $\mathcal{L}_{\mathrm{test}}$ when $b=1$, we plot the best result in terms of MISE during 40 trials in Figure \ref{Fig:BadB}. We see that GP4C ($b=1$) over-estimates the variance of the intensity function and the over-estimated variance leads to the poor performance in $\mathcal{L}_{\mathrm{test}}$. We fix $b=0.3$ during the remaining experiments for simplicity.

\begin{figure*}[ht]
	\includegraphics[width=\columnwidth]{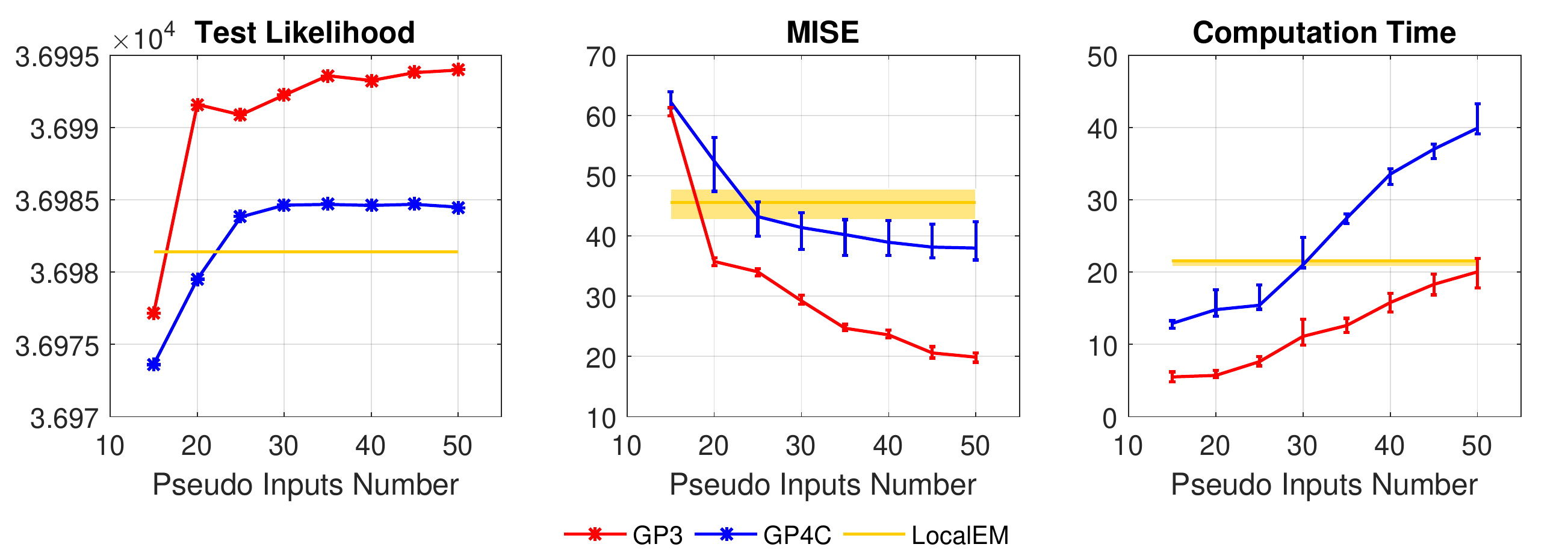}
	\caption{\textbf{Synthetic Data Set}. Comparison of performance of GP3, GP4C and LocalEM in terms of $\mathcal{L}_\mathrm{test}$, MISE and $T$ when varying the number of pseudo inputs for sparse GPs. For MISE and the computation time, the median, the 0.25 and 0.75 quantiles of the statistics in 40 experiments are shown with error bars or shaded area. For GP3 and GP4C, MISE and $\mathcal{L}_\mathrm{test}$ stay relatively stable with the increase of the number of pseudo inputs.}
	\label{Fig:SynAP}
\end{figure*}

\textbf{Number of the pseudo inputs}. We vary the number of pseudo inputs in the GP3 and GP4C since this number determines the accuracy of approximation in a sparse GP. We expect for GP-based methods, the test likelihood will be relatively stable when we increase the number of pseudo inputs according to previous studies on Sparse GPs \cite{titsias2009variational}. 

The result for the Synthetic A data set is given in Figures \ref{Fig:SynAP}. In Figure \ref{Fig:SynAP}, we see that for GP3 and GP4C, MISE and $\mathcal{L}_\mathrm{test}$ stay relatively stable with the increase of the number of pseudo inputs. The computation time of GP3 and GP4C will grow with the increase of the number of pseudo inputs. 

In both Table \ref{Tab:Choicesb} and Figure \ref{Fig:SynAP}, We see that GP4C outperforms LocalEM on these three datasets. However, we also notice that there is still a gap between GP3 and GP4C in terms of $\mathcal{L}_{\mathrm{test}}$ and MISE in Table \ref{Tab:Choicesb}. Two reasons may account for this fact. The first one is that the data are provided in the form of panel counts rather than exact timestamps. The second reason is that we use a lower bound of the true ELBO to perform the variational inference, which may lead to a bias. This bias can be alleviated with the stochastic variational inference \cite{paisley2012variational}, where our lower bound can serve as a control variate. We leave this as a future study.

An additional experiment in which we increase the number of training subjects to evaluate the gain in performance on Synthetic A data set is given in Appendix E.

\subsection{REAL WORLD DATA SETS}

\begin{table}[t]
	\caption{Statistics about the three data sets, where $K$, $\mathcal{X}$, $\bar{N}$ and $N$ denote the number of subjects in each data sets, the underlying continuous space, the number of different end points and the number of different intervals $\mathcal{X}_i^{(k)}$, respectively.}
	\label{Table:StatReal}
	\begin{center}
		\begin{tabular}{lllll}
			\multicolumn{1}{c}{\textbf{Data Set}} &\multicolumn{1}{c}{$\mathcal{X}$} &\multicolumn{1}{c}{$K$}  &\multicolumn{1}{c}{$\bar{N}$}& \multicolumn{1}{c}{$N$}\\
			\hline \\[-1.5ex]
			Nausea (A) & $[0,55]$ & 65 & 45 & 109\\
			Nausea (B) & $[0,55]$ & 48 & 38 & 84\\
			\hline \\[-1.5ex]
			Bladder (A) & $[0,53]$ &38 & 52 & 176\\
			Bladder (B) & $[0,53]$ &47 & 52 & 201\\
			\hline \\[-1.5ex]
			Skin (A\&B) & $[0,61.57]$ &143 & 751 & 816\\
			Skin (C\&D) & $[0,62.63]$ &147 & 808 & 887\\
		\end{tabular}
	\end{center}
\end{table}

\citet{sun2016statistical} provided three panel count data sets. Some statistics can be found in Table \ref{Table:StatReal}. We see that among the three data sets, the Nausea and Bladder sets are smaller in terms of the number of subjects. A brief description about the three data sets is as follows. We use 18 pseudo inputs for all real world experiments. In the $s$th trial, we randomly partition each data set into two parts, which are $\mathcal{D}_{\mathrm{train}}^{(s)}$ and $\mathcal{D}_{\mathrm{test}}^{(s)}$.

\begin{itemize}
	\item [a)] \textbf{Nausea data set}. This data set contains the visiting times from 113 patients during 52 weeks. The panel count data were obtained by recording the reported count of vomits from each patient between two subsequent visits. Patients were divided into two groups, which are the treatment group (65 patients) and the placebo group (48 patients). We denote the two groups by the Nausea A (treatment) and B (placebo) set.
	\item [b)] \textbf{Bladder cancer data set}. This data set arises from a bladder cancer study conducted by the Veterans Administration Cooperative Urological Research Group. It records the counts of new bladder tumors that occurred between subsequent visits from 85 patients during 53 weeks, who were divided into the placebo group (47 patients) and the treatment group (38 patients). We denote the two groups as the Bladder A (treatment) and B (placebo) set, respectively.
	\item [c)] \textbf{Skin cancer data set}. This data set were recorded during a skin cancer experiment conducted by the University of Wisconsin Comprehensive Cancer Center and the numbers of new skin cancers of two different types between two subsequent visits from 290 patients were recorded during five years. The visiting time was recorded in the form of days since the first visit and we divided the days by 30. Patients were divided into treatment and placebo groups. Let the panel count data in treatment group be Skin A and B sets and the panel count data in placebo group be Skin C and D sets. 
\end{itemize}

On these three data sets, since the original data are in the form of panel counts, GP3 is not used as a comparison. We compare GP4C with the localEM method in terms of $\mathcal{L}_{\mathrm{test}}$ and the computation time $T$. The results are given in Table \ref{Table:ResultReal}. LocalEM performs better on the Nausea and Bladder data sets in terms of the computation time $T$. In all data sets, GP4C performs well on $\mathcal{L}_{\mathrm{test}}$ and outperforms LocalEM on computation time on the Skin data sets.

\begin{wrapfigure}{l}{8cm}
	\includegraphics[width=0.5\columnwidth]{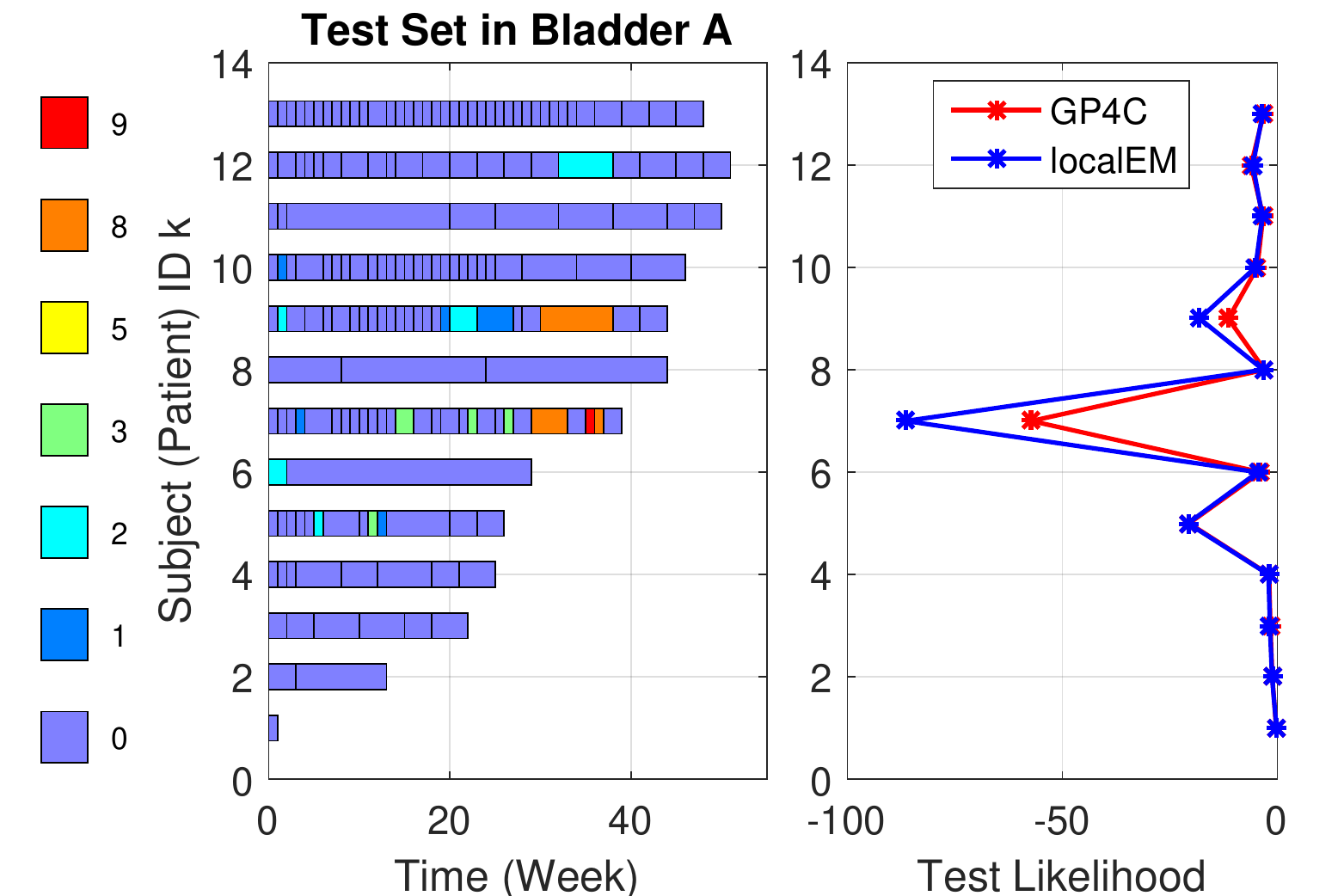}
	\caption{\textbf{Bladder A Data Set}. An illustration of the panel count data in the test set (Left) and the test likelihood from GP4C and LocalEM of each subject (Right). GP4C mainly outperforms LocalEM on two subjects whose numbers of newly-occurred cancers are large (No. 7 and 9).}
	\label{Fig:TestBladder}
\end{wrapfigure}

\begin{table}[t]
	
	\begin{center}
		\begin{tabular}{lllll}
			\multicolumn{1}{c}{\textbf{Data Set}} &\multicolumn{1}{c}{\textbf{METHOD}}  & \multicolumn{1}{c}{ $T [s]$}&\multicolumn{1}{c}{\bf $\mathcal{L}_\mathrm{test}$}\\
			\hline \\[-1.5ex]
			Nausea (A) 
			& localEM & 0.98           & -156.99\\
			& GP4C    & 11.42           & \textbf{-155.97}\\
			Nausea (B) 
			& localEM & 0.71           &  -183.61 \\
			& GP4C    & 10.10           & \textbf{-104.94}\\
			\hline \\[-1.5ex]
			Bladder (A) 
			& localEM & 1.15           & -122.11\\
			& GP4C    & 29.63          & \textbf{-107.82}\\
			Bladder (B) 
			& localEM & 1.12           & -147.61\\
			& GP4C    & 20.32          & \textbf{-146.23}\\
			\hline \\[-1.5ex]
			Skin (A) 
			& localEM & 62.88          & -161.86\\
			& GP4C    & 34.96          & \textbf{-161.48}\\
			Skin (B) 
			& localEM & 62.67          & -121.12\\
			& GP4C    & 34.57          & \textbf{-117.48}\\
			Skin (C) 
			& localEM & 74.07          & -228.48\\
			& GP4C    & 19.69          & \textbf{-227.22}\\
			Skin (D) 
			& localEM & 72.77          & -128.47\\
			& GP4C    & 34.64          & \textbf{-128.24}\\
		\end{tabular}
	\end{center}
\caption{The comparison of the test likelihood ($\mathcal{L}_\mathrm{test}$) and the computation time $T$ measured in seconds on the three panel count data sets. LocalEM performs better on the Nausea and Bladder data sets in terms of computation time. In all data sets, GP4C performs well on the test Likelihood and performs well on computation time on the Skin data sets.}
\label{Table:ResultReal}
\end{table} 

To see the difference between GP4C and LocalEM, we show the result of inferred intensity by two algorithms during one trial on the Bladder A data set in Figure \ref{Fig:DemoIntensity}. We see that GP4C provides the additional uncertainty which helps improve $\mathcal{L}_{\mathrm{test}}$ compared with LocalEM. Since the Bladder A set is small, we plot the panel count data in the training set in Figure \ref{Fig:DemoPanel}. The test set and the test likelihood of all its subjects are given in Figure \ref{Fig:TestBladder}. From the test likelihood of each subject, we see that GP4C outperforms LocalEM on two subjects whose counts of newly-occurred tumors are large (No. 7 and No. 9). The count 8 never occurs in the training set and a point estimate will fail to model this uncertainty while a GP-modulated method will take the uncertainty into consideration. 

Another observation about this data set is that there is a heterogeneity across all subjects and we can add an additional variable on the intensity function for each subject to describe the unobservable independent random effects \cite{cook2007statistical}. We briefly discuss how to add this change to GP4C and conduct experiments on real world data sets in Appendix F.\\


\section{CONCLUSION}
We presented the first framework for GP-modulated Poisson processes when data appear in the form of panel count data. We derived a tractable lower bound for the intractable evidence lower bound when modeling the panel count data using the GP-modulated intensity function. In the future, we plan to implement the stochastic variational inference algorithm to evaluate the bias in the tractable lower bound. We are also considering to find an applicable two-dimensional data set where we can extend our algorithm to spatial point processes.

\bibliography{uai2018.bib}

\newpage
\appendix
\section{DETAILS ON LEMMA 1}
Before we proceed, we state a technical result:
\begin{lemma}
	Let $y\sim \mathcal{N}(\mu,\sigma^2)$ and $\varphi = (\mu/\sigma)^2$. Then 
	\begin{equation}
	\mathbb{E}_y[\ln y^2] = \ln(2\sigma^2)+\sum_{j=0}^{\infty} \frac{(\varphi/2)^j\exp(-\varphi/2)}{j!}\psi(j+1/2),
	\end{equation}
	where $\psi(\cdot)$ is the digamma function.
	\label{app:lem:1}	
	\begin{proof}
		Let $\tilde{y}=y/\sigma$, then the expectation can be calculated as
		\begin{equation}
		\mathbb{E}_y[\ln y^2]=\int_{-\infty}^{\infty}\ln y^2\frac{1}{\sqrt{2\pi}\sigma}\exp\Big(-\frac{(y-\mu)^2}{2\sigma^2}\Big)dy = \ln\sigma^2 + \int_{-\infty}^{\infty}\ln \tilde{y}^2\frac{1}{\sqrt{2\pi}}\exp\Big(-\frac{(\tilde{y}-\mu/\sigma)^2}{2}\Big) d\tilde{y}.
		\end{equation}
		The second part has the form of $\mathbb{E}_{\bar{y}}[\ln\bar{y}^2]$, where $\bar{y}\sim \mathcal{N}(\mu/\sigma,1)$ . Let $w=\bar{y}^2$ and $w$ follows a standard non-central chi-squared distribution with parameter $\varphi = (\mu/\sigma)^2$ \cite{famoye1995continuous}. The distribution of $w$ is given as follows:
		\begin{equation}
		p(w)=\frac{e^{-\frac{w+\varphi}{2}}}{\sqrt{2w}}\sum_{j=0}^{\infty}\frac{(w\varphi/4)^j}{j!\Gamma(j+1/2)}.
		\end{equation}
		The expectation of $\ln w$ then is
		\begin{equation}
		\mathbb{E}_w[\ln w]= \int_{0}^{\infty} \ln w\frac{e^{-\frac{w+\varphi}{2}}}{\sqrt{2w}}\sum_{j=0}^{\infty}\frac{(w\varphi/4)^j}{j!\Gamma(j+1/2)}dw = \sum_{j=0}^{\infty} \frac{(\varphi/2)^je^{-\varphi/2}}{j!}(\ln 2+\psi(j+1/2)). 
		\end{equation}
		Substituting this back yields the answer.
	\end{proof}
\end{lemma}

\section{DETAILS ON LEMMA 2}
Let us recall that
\begin{equation}
g_m(x) = \sum_{j=0}^{\infty} \frac{x^j\exp(-x)}{j!}\psi(j+m). 
\end{equation}
The derivative of $g_m(x)$ with respect to $x$ is
\begin{equation}
g_m^\prime(x) = \sum_{j=0}^{\infty} \frac{(jx^{j-1}-x^j)\exp(-x)}{j!}\psi(j+m)= \sum_{j=0}^{\infty} \frac{x^j\exp(-x)}{j!}\frac{1}{j+m}.
\end{equation}
To prove the Lemma 2 in Section 4, we first present two results:
\begin{lemma}{\cite{moser2007some}}
	\begin{align}
	&g_m^\prime(x)\geq \frac{1}{x+m},~m\in \mathbb{N}^+,x>0.\nonumber
	\end{align} \label{app:lem:2}
\end{lemma}
Note that the inequality holds when $m\in\mathbb{N}_+$. However, following the same lines of the proof, one can generalized their results for $m\in \mathbb{R}^+$, hence the proof is elided.
In our case, we are interested in a bound when $m=\frac{1}{2}.$ We state the following:
\begin{lemma}
	The following inequality holds:
	\begin{equation}
	g_m(x) \geq \ln(x+m)+\psi(m)-\ln(m).
	\end{equation}
	\begin{proof}
		Since 
		\begin{equation}
		\frac{1}{x+m}\leq g_m^\prime(x),
		\end{equation}
		integrating both sides yield
		\begin{equation*}
		\ln(x+m)-\ln m = \int_{0}^{x}\frac{1}{y+m}dy\leq\int_{0}^{x}g_m^\prime(y)dy=g_m(x)-g_m(0)=g_m(x)-\psi(m).
		\end{equation*}
	\end{proof}\label{app:lem:3}
\end{lemma}
\begin{lemma}
	Let $x\sim \mathcal{N}(\mu,\sigma^2)$. Then we have
	\begin{equation}
	\mathbb{E}_x[\ln x^2]\geq \ln(\mu^2+b\sigma^2)-C-\ln 2,~b\in[0,1],
	\end{equation}
	where $C$ is Euler's constant and takes the value $\approx 0.5772$.
	\begin{proof}
		Invoking Lemma \ref{app:lem:2}, it is obvious that the inequality holds true for $b=1$,
		\begin{align}
		&\mathbb{E}_x[\ln x^2]  = \ln(2\sigma^2)+g_{0.5}\Big(\frac{\mu^2}{2\sigma^2}\Big)\nonumber\\
		&\geq \ln(2\sigma^2)+\ln\Big(\frac{\mu^2}{2\sigma^2}+\frac{1}{2}\Big)+\psi(1/2)+\ln(2)= \ln(\mu^2+\sigma^2)-C-\ln 2.
		\end{align}
		This implies that the inequality holds true for all values of $b\in[0,1]$. 
	\end{proof}
	\label{app:lem:4}
\end{lemma}

\section{DETAILS ON THEOREM 2}
Theorem 2 can be obtained by applying Theorem 1 on the ELBO $\mathcal{L}$. In Theorem 2, there are two expectations $\mathbb{E}_q^2f(x)$ and $\mathrm{Var}_qf(x)$ which can be 
computed as follows \cite{lloyd2015variational}:

\begin{align}
\mathbb{E}_q^2f(x)&=\mathrm{tr}(K_{RR}^{-1}\Phi K_{RR}^{-1}(\bm{\mu\mu}^\top)),\\
\mathrm{Var}_qf(x)&=\gamma|\mathcal{X}^{(k)}|-\mathrm{tr}(K_{RR}^{-1}\Phi)+\mathrm{tr}(K_{RR}^{-1}\Phi K_{RR}^{-1}\Sigma).
\end{align}

\section{TEST LIKELIHOOD OF GP4C and GP3}
Recall that during the $s$th trial, the test likelihood is
\begin{align}
\mathcal{L}_\mathrm{test}(s) &\stackrel{\Delta}{=} \ln \int p(\mathcal{D}_{\mathrm{test}}^{(s)}|f)p(f|\mathcal{D}_\mathrm{train}^{(s)}) df\nonumber\\
&\approx\ln \frac{1}{U}\sum_{u=1}^{U} p(\mathcal{D}_{\mathrm{test}}^{(s)}|f^{(s,u)})\label{app:equ:sampleU}\\
&= \ln \sum_{u=1}^{U}\exp\Big(\ln p(\mathcal{D}_{\mathrm{test}}^{(s)}|f^{(s,u)})\Big)-\ln U\nonumber\\
&= \ln \sum_{u=1}^{U}\exp\Big(\sum_{k=1}^{K_\mathrm{test}}\sum_{i=1}^{N_k} \Big(m_i^{(k)}\ln r_{ik}^{(s,u)}-\ln (m_i^{(k)}!)\Big) -\sum_{k=1}^{K_\mathrm{test}}\int_{\mathcal{X}^{(k)}}\Big(f^{(s,u)}(x)\Big)^2dx\Big)-\ln U.\label{app:equ:omit}
\end{align}
In the above derivation, we use
\begin{align}
&f^{(s,u)}\sim \mathcal{N}(\mu^{(s)},\Sigma^{(s)}),\\
&r_{ik}^{(s,u)}=\int_{\mathcal{X}^{(k)}_i}\Big(f^{(s,u)}(x)\Big)^2dx.
\end{align}

We can also calculate the test likelihood for each subject similarly. In Equation \eqref{app:equ:sampleU}, we draw $U=50$ samples of the function $f^{(s,u)}$ from the variational distribution $q^{(s)}(f)$ on a vector of 3001 evenly-spaced points on $\mathcal{X}$ and we approximate points at an arbitrary position on $\mathcal{X}$ with the linear interpolation. The log-exp-sum trick is used to calculate the $\mathcal{L}_\mathrm{test}(s)$. We calculate all integrals in $p(\mathcal{D}_\mathrm{test}^{(s)}|f)$ using Simpson's rule with 501 evenly-spaced points. 

In Equation \eqref{app:equ:omit}, the term $\sum_{k}\sum_i\ln(m_i^{(k)}!)$ can be extracted out and treated as a constant.

\section{ADDITIONAL SYNTHETIC EXPERIMENTS}
\textbf{Ratio of training subjects}. We vary the number of training subjects by adjusting the ratio relative to full training subjects. We expect all methods will benefit from the increase of the training subjects.

\begin{figure*}[ht]
	\includegraphics[width=\columnwidth]{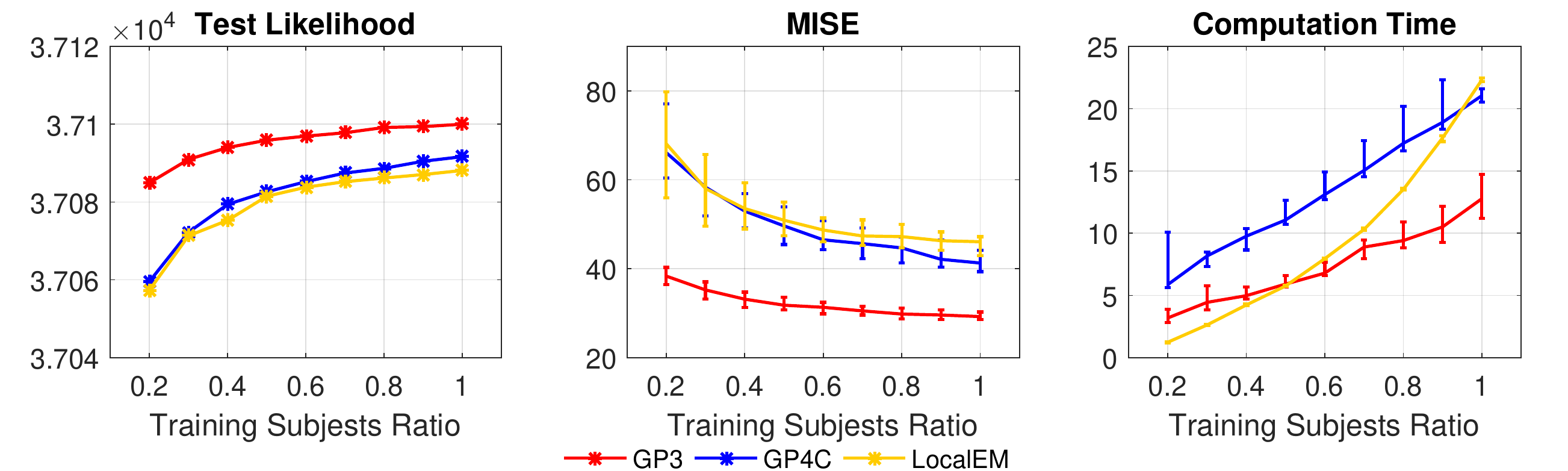}
	\caption{\textbf{Synthetic Data Set}. Comparison of performance of GP3, GP4C and LocalEM in terms of $\mathcal{L}_\mathrm{test}$, MISE and $T$ when varying the ratio of training subjects and the test set is the same. For MISE and the computation time, the 0.25 and 0.75 quantiles of the statistics in 40 experiments are shown with error bars. All methods benefit from the increase of the number of training subjects. The computation time of GP3 and GP4C grow linearly with the increase of the number of training subjects.}
	\label{Fig:SynAR}
\end{figure*}

The result for the Synthetic A data set is given in Figure \ref{Fig:SynAR}. We see that all three methods benefit from the increase of the number of training subjects. The computation time of GP3 and GP4C grow linearly with the increase of the number of training subjects but LocalEM grows more rapidly.

\section{GP4C MODEL WITH INDIVIDUAL WEIGHT}
\begin{figure}
	\begin{subfigure}{.5\textwidth}
		\includegraphics[width=\columnwidth]{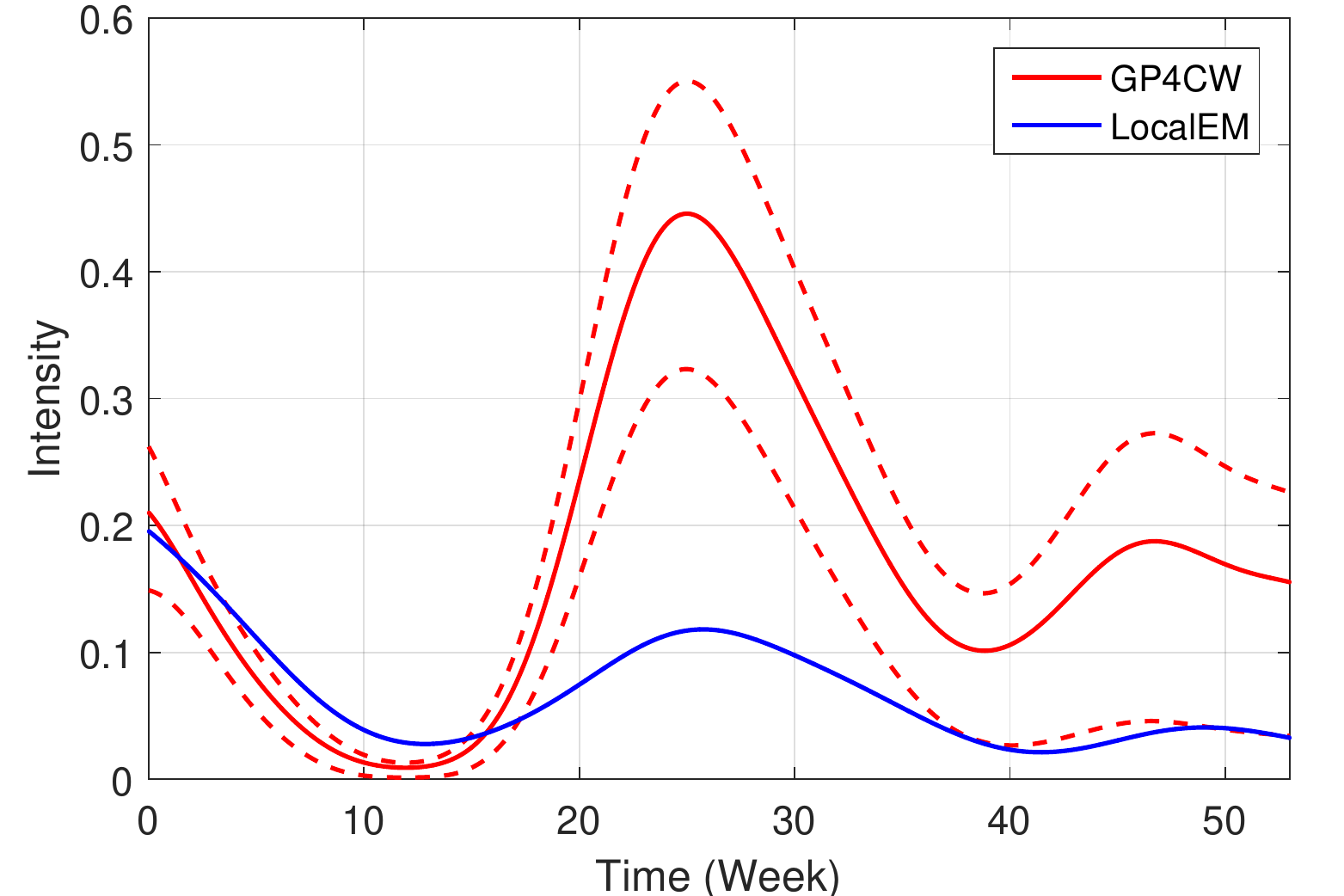}
		\caption{\textbf{Bladder A Data Set}.}
		\label{Fig:DemoIntensityW}
	\end{subfigure}%
	\begin{subfigure}{.5\textwidth}
		\includegraphics[width=\columnwidth]{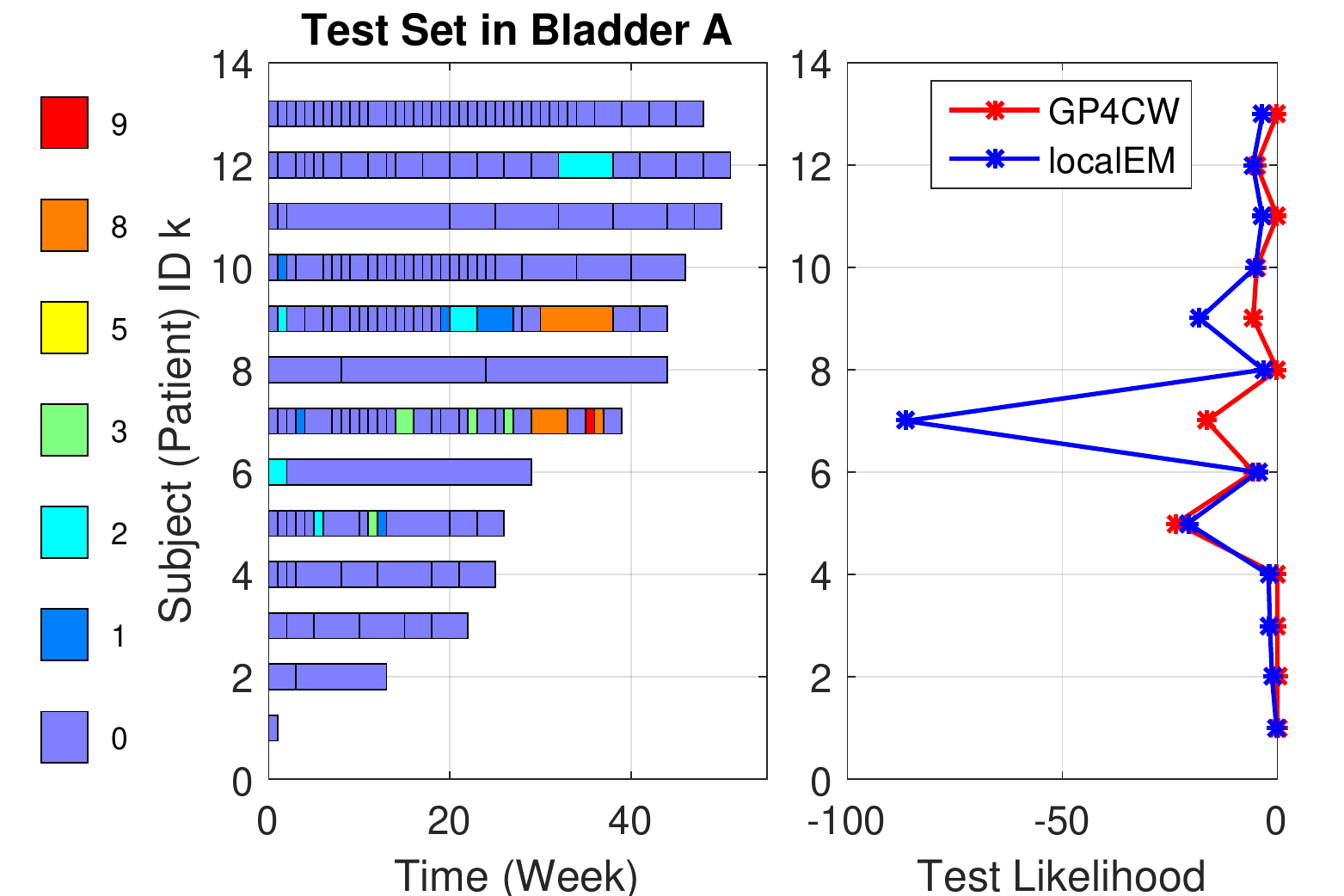}
		\caption{\textbf{Bladder A Data Set}.}
		\label{Fig:TestBladderW}
	\end{subfigure}
	\caption{(a) Inferred intensity function by the LocalEM and GP4CW methods. For GP4CW, a 75\% credible interval is given by dotted lines. (b) An illustration of the panel count data in the test set (Left) and the test likelihood from GP4C and LocalEM of each subject (Right). GP4CW mainly outperforms LocalEM on two subjects whose numbers of newly-occurred cancers are large (No. 7 and 9).}
\end{figure}

\subsection{MODEL}
It is practical to assume that the $k$'th subject has an individual weight parameter $\upsilon_k$ multiplied to the basic intensity function, because in traditional panel count data sets, each subject is a patient whose personal information, such as age, is not the same and the count data from each patient may vary greatly. Such a modification is called the unobservable independent random effects in \citet{cook2007statistical}. In the simplest case, we consider the following model for the underlying intensity function:
\begin{equation}
\lambda_k(x) = \upsilon_k f^2(x),~f\sim \mathcal{GP}(g(x),\kappa(x,x')),
\label{Equ:VarModel}
\end{equation}
where $\upsilon_k\in \mathbb{R}^+$ is a deterministic and positive real number. The likelihood is as follows.
\begin{equation}
p(\mathcal{D},f) = \Big[\prod_{k=1}^{K}p(\bm{d}_k|\lambda(x);\upsilon_k)\Big]p(f;g,\kappa).
\end{equation}

We call this model \textbf{GP4C} model with individual \textbf{W}eight (GP4CW). 

We can further generalize this model by assuming that the intensity function of the $k$'th subject is a linear combination of basis intensity functions \cite{lloyd2016latent} and the mixture weights are also deterministic. 

\subsection{INFERENCE}
The inference of GP4CW is almost the same as GP4C. We only need to modify GP4C by adding the inference of the point estimate of $\upsilon_k$ in M-step of the vEM framework as follows.
\begin{equation}
\upsilon_k = \max\Big\{\epsilon,\frac{\sum_{i=1}^{N_k}m_i^{(k)}}{\int_{\mathcal{X}^{(k)}}\mathbb{E}_q[f^2(x)]dx}\Big\},
\end{equation}
where $\epsilon=10^{-6}$ is a small number to guarantee the positiveness of $\upsilon_k$.

\subsection{EXPERIMENT ON THE REAL WORLD DATA SET}
On the three real world data sets. The test likelihood $\mathcal{L}_{\mathrm{test}}$ and the computation time $T$ are given in Table \ref{app:Table:ResultRealFull}. We also plot the test likelihood of each subject and the inferred intensity function from GP4CW in Figures \ref{Fig:TestBladderW} and \ref{Fig:DemoIntensityW}. We can notice that GP4CW provides more accurate estimation on the patient No. 7 and No. 9.

\begin{table}[h]
	\caption{The comparison of the test likelihood ($\mathcal{L}_\mathrm{test}$) and the computation time ($T$) on the three panel count data sets for GP4C, GP4CW and LocalEM.}
	\label{app:Table:ResultRealFull}
	\begin{center}
		\begin{tabular}{lllll}
			\multicolumn{1}{c}{\textbf{Data Set}} &\multicolumn{1}{c}{\textbf{METHOD}}  & \multicolumn{1}{c}{ $T[s]$}&\multicolumn{1}{c}{\bf $\mathcal{L}_\mathrm{test}$}\\
			\hline \\[-1.5ex]
			Nausea (A) 
			& localEM & 0.98           & -156.99\\
			& GP4C    & 11.42           & -155.97\\
			& GP4CW   & 11.95          & \textbf{65.23}\\
			Nausea (B) 
			& localEM & 0.71           &  -183.61 \\
			& GP4C    & 10.10           & -104.94\\
			& GP4CW   & 19.22          & \textbf{49.81}\\
			\hline \\[-1.5ex]
			Bladder (A) 
			& localEM & 1.15           & -122.11\\
			& GP4C    & 29.63          & -107.82\\
			& GP4CW   & 38.74          & \textbf{-40.80}\\
			Bladder (B) 
			& localEM & 1.12           & -147.61\\
			& GP4C    & 20.32          & -146.23\\
			& GP4CW   & 38.89          & \textbf{-74.87}\\
			\hline \\[-1.5ex]
			Skin (A) 
			& localEM & 62.88          & -161.86\\
			& GP4C    & 34.96          & -161.48\\
			& GP4CW   & 31.91          & \textbf{-111.90}\\
			Skin (B) 
			& localEM & 62.67          & -121.12\\
			& GP4C    & 34.57          & -117.48\\
			& GP4CW   & 29.66	       & \textbf{-62.65}\\
			Skin (C) 
			& localEM & 74.07          & -228.48\\
			& GP4C    & 19.69          & -227.22\\
			& GP4CW   & 18.86          & \textbf{-150.54}\\
			Skin (D) 
			& localEM & 72.77          & -128.47\\
			& GP4C    & 34.64          & -128.24\\
			& GP4CW   & 39.12          & \textbf{-78.39}\\
		\end{tabular}
	\end{center}
\end{table} 

\end{document}